\documentclass[sigconf]{aamas}  % do not change this line!
\usepackage{balance}  % do not change this line -- unless you manually balance your last page

\usepackage{booktabs}

\setcopyright{ifaamas}  % do not change this line!
\acmDOI{}  % do not change this line!
\acmISBN{}  % do not change this line!
\acmConference[AAMAS'19]{Proc.\@ of the 18th International Conference on Autonomous Agents and Multiagent Systems (AAMAS 2019)}{May 13--17, 2019}{Montreal, Canada}{N.~Agmon, M.~E.~Taylor, E.~Elkind, M.~Veloso (eds.)}  % do not change this line!
\acmYear{2019}  % do not change this line!
\copyrightyear{2019}  % do not change this line!
\acmPrice{}  % do not change this line!

\begin{document}

\title{Malthusian Reinforcement Learning}  % put your title here!
%\titlenote{Titlenote --- e.g., used for things like "This article extends an earlier paper titled XYZ", and "equal contribution by the first two authors".}

% AAMAS: as appropriate, uncomment one subtitle line; see camera ready instructions
%\subtitle{Extended Abstract}
%\subtitle{Blue Sky Ideas Track}
%\subtitle{JAAMAS Track}
%\subtitle{Doctoral Consortium}                              
%\subtitle{Demonstration}
%\subtitlenote{Please refrain from using subtitle notes}

\author{Joel Z. Leibo}
%\authornote{lorum ipsum}
\orcid{0000-0002-3153-916X}
\affiliation{%
 \institution{DeepMind}
 \streetaddress{6 Pancras Square}
 \city{London} 
 \state{UK} 
 \postcode{N1C 4AG}
}
\email{jzl@google.com}
\author{Julien Perolat}
%\authornote{lorum ipsum}
\affiliation{%
 \institution{DeepMind}
 \streetaddress{6 Pancras Square}
 \city{London} 
 \state{UK} 
 \postcode{N1C 4AG}
}
\email{perolat@google.com}
\author{Edward Hughes}
%\authornote{lorum ipsum}
\affiliation{%
 \institution{DeepMind}
 \streetaddress{6 Pancras Square}
 \city{London} 
 \state{UK} 
 \postcode{N1C 4AG}
}
\email{edwardhughes@google.com}
\author{Steven Wheelwright}
%\authornote{lorum ipsum}
\affiliation{%
 \institution{DeepMind}
 \streetaddress{6 Pancras Square}
 \city{London} 
 \state{UK} 
 \postcode{N1C 4AG}
}
\email{sjwheel@google.com}
\author{Adam H. Marblestone}
%\authornote{lorum ipsum}
\affiliation{%
 \institution{DeepMind}
 \streetaddress{6 Pancras Square}
 \city{London} 
 \state{UK} 
 \postcode{N1C 4AG}
}
\email{amarblestone@google.com}
\author{Edgar Du\'e\~nez-Guzm\'an}
%\authornote{lorum ipsum}
\affiliation{%
 \institution{DeepMind}
 \streetaddress{6 Pancras Square}
 \city{London} 
 \state{UK} 
 \postcode{N1C 4AG}
}
\email{duenez@google.com}
\author{Peter Sunehag}
%\authornote{lorum ipsum}
\affiliation{%
 \institution{DeepMind}
 \streetaddress{6 Pancras Square}
 \city{London} 
 \state{UK} 
 \postcode{N1C 4AG}
}
\email{sunehag@google.com}
\author{Iain Dunning}
%\authornote{lorum ipsum}
\affiliation{%
 \institution{DeepMind}
 \streetaddress{6 Pancras Square}
 \city{London} 
 \state{UK} 
 \postcode{N1C 4AG}
}
\email{idunning@google.com}
\author{Thore Graepel}
%\authornote{lorum ipsum}
\affiliation{%
 \institution{DeepMind}
 \streetaddress{6 Pancras Square}
 \city{London} 
 \state{UK} 
 \postcode{N1C 4AG}
}
\email{thore@google.com}

\renewcommand{\shortauthors}{JZ Leibo et al.}

\begin{abstract}
Here we explore a new algorithmic framework for multi-agent reinforcement learning, called Malthusian reinforcement learning, which extends self-play to include fitness-linked population size dynamics that drive ongoing innovation. In Malthusian RL, increases in a subpopulation's average return drive subsequent increases in its size, just as Thomas Malthus argued in 1798 was the relationship between preindustrial income levels and population growth \cite{malthus1798essay}. Malthusian reinforcement learning harnesses the competitive pressures arising from growing and shrinking population size to drive agents to explore regions of state and policy spaces that they could not otherwise reach. Furthermore, in environments where there are potential gains from specialization and division of labor, we show that Malthusian reinforcement learning is better positioned to take advantage of such synergies than algorithms based on self-play.
\end{abstract}

  % put your semicolon-separated keywords here!

\maketitle

%%%%%%%%%%%%%%%%%%%%%%%%%%%%%%%%%%%%%%%%%%%%%%%%%%%%%%%%%%%%%%%%%%%%%%%%%%%%%%%%%%%%%%%%%%%%%%%%%%%%%%%%%
%% start of main body of paper

\section{Introduction}

Reinforcement learning algorithms have considerable difficulty avoiding local optima and continually exploring large state and policy spaces. This is known as the problem of exploration. In single-agent reinforcement learning, the main approach is to rely on intrinsic motivations, e.g., for individual curiosity \cite{schmidhuber2010formal, bellemare2016unifying, ostrovski2017count, pathak2017curiosity, martin2017count, burda2018large}, empowerment \cite{klyubin2005empowerment}, or social influence \cite{jaques2018intrinsic}.

However, critical adaptation events in human history are difficult to explain with intrinsic motivations. Consider the dispersal of homo sapiens out of Africa, where they first evolved between $200,000$ and $150,000$ years ago, throughout the globe, eventually occupying essentially all terrestrial climatic conditions and habitats by $14,500$ years before the present \cite{goebel2008late}. This example is relevant to AI research because intelligence is often defined as an ability to adapt to a diverse set of environments \cite{legg2007universal}. Essentially no other process on earth has led so quickly to so much adaptive diversity as did the expansion of human foragers throughout the globe\footnote{The rapid dispersal of homo sapiens across the globe and their adaptation to the full range of diverse terrestrial habitats can be regarded as a great feat of intelligence. We may even assess its universal intelligence $\Upsilon$ by adapting the following definition from \cite{legg2007universal}, $\Upsilon(\pi) = \sum_{h \in \mathcal{H}} 2^{-K(h)} V_h^\pi$, where $\pi$ is a policy, $H$ is the set of terrestrial habitats, $V_h^\pi$ is the expected value of the sum of rewards from inhabiting the habitat $h$, and $K(h)$ is a measure of the complexity of $h$. The complexity measure could be defined as any sensible ecological distance metric measuring the distance from the ancestrally adapted environment to $h$.}. Human foraging communities were capable both of discovering water-finding strategies suitable for the arid Australian desert as well as how to hunt seals hiding beneath ice sheets and keep warm in the Arctic \cite{boyd2011cultural}. From this perspective, the great dispersal of human foragers can be seen as some of the best evidence for an ``existence proof'' that intelligence, by this definition, is even possible. Yet there's no evidence that intrinsic motivations like curiosity played a role in it. Rather, considerable evidence points to a variety of extrinsic motivation mechanisms as the main drivers of human migration including climate change \cite{stewart2012human, eriksson2012late} and demographic expansion \cite{mellars2006did, powell2009late}.
    
One perspective on the problem of exploration is that the difficulty comes from the sparseness of extrinsic rewards. If extrinsic rewards are very sparse, then it is hard to estimate state-value functions and policy gradients since returns will have very high variance. Thus intrinsic motivation methods produce frequent intermediate (dense) rewards in hopes of bridging the long gaps between extrinsic rewards \cite{chentanez2005intrinsically}. Multi-agent reinforcement learning offers an alternative. Algorithms based on self-play like AlphaGo \cite{tesauro1995td, Silver16Go, silver2017chess, bansal2017emergent, jaderberg2018human} are aimed at an essentially single-agent objective, e.g., defeat a specific human grandmaster. Directly training by single-agent reinforcement learning to accomplish that objective is a lost cause. Since the untrained agent would never win a game, it would never get any reward signal to learn from. Self-play, on the other hand, provides an alternative incentive for agents to explore deeply through the strategy space. In two-player zero-sum, it escapes local optima by learning to exploit them, diminishing the returns available from such strategies, and thereby extrinsically motivating new exploration. Of course, life is not a zero-sum game. Nevertheless, real-life feats of exploration like the dispersal of homo sapiens out of Africa really were motivated in part by a pressure to out-compete rivals in a struggle for scarce resources that became increasingly difficult over time due to demographic expansion. Moreover, local competition between individuals is a universal feature of natural habitats, and underlies the evolution of dispersal \cite{doi:10.1146/annurev.es.21.110190.002313}. In this paper we investigate whether such population size dynamics can be exploited as an algorithmic mechanism in multi-agent reinforcement learning.
    
As an algorithmic mechanism, augmenting multi-agent reinforcement learning with population size dynamics appears to have the requisite property needed to evade local optima and traverse large state spaces. Strategies that work well at low densities do not necessarily translate well to high densities, but success at any density ensures density will increase further in the future. The rules of the game therefore naturally shift over time in a manner that depends on past outcomes. This ensures that species cannot remain too long in comfortable local optima. When resources are scarce, rising populations eventually dissipate gains from learning, forcing agents to innovate just to maintain existing reward levels \cite{clark2008farewell}\footnote{\cite{clark2008farewell} argues that preindustrial human populations generally oscillated around a fixed, and only very slowly increasing, carrying capacity until the industrial revolution. Similar oscillations in subpopulation sizes were recently observed in a large-scale multi-agent learning simulation by \cite{yang2018study}. As those authors pointed out, its possible for population dynamics to endlessly oscillate rather than increasing over time. The same is true for the strategies used by learning algorithms based on self-play. One fix that was used in AlphaGo and elsewhere is to require agents to learn to defeat all previous versions of themselves, not just the most recent \cite{Silver16Go, silver2017chess}. This prevents self-play-based agents from endlessly learning and forgetting the same exploit and defense.}.
    
So far we've motivated introducing population dynamics to multi-agent reinforcement learning by appealing to a  competitive struggle for existence against a well-matched foe, i.e., the same argument underlying the performance of self-play in two-player zero-sum games. However, there is more to life than competition. As an algorithmic mechanism to promote learning in general-sum multi-agent environments, population dynamics may also be more suitable than self-play-based approaches. In particular, we will consider whether this approach provides greater scope for adapting to synergies between specialists, making it easier to discover joint policies involving significant division of labor.

In this work we introduce a new algorithm for multi-agent reinforcement learning based on these principles of population dynamics. It is called Malthusian reinforcement learning because improvements in returns for any subpopulation translate directly into increases in the size of that subpopulation in subsequent episodes. Thus it may be evaluated on either the individual or the group level. In this work we are interested in two specific questions:
\begin{enumerate}
    \item Is Malthusian reinforcement learning better at avoiding becoming stuck in bad local optima in \emph{individual} policy space than competing algorithms based on intrinsic motivation?
    
    \item Is it easier to evolve joint policies to implement heterogeneous mutualism behaviors with Malthusian reinforcement learning than with alternative approaches based on self-play?
\end{enumerate}

\section{Model}

\subsection{Characteristics of the Malthusian reinforcement learning framework}

Malthusian reinforcement learning differs from standard multi-agent reinforcement paradigms in a number of ways.

\begin{enumerate}
    \item Malthusian reinforcement learning may be seen as an algorithm for ``community coevolution''. It produces a set of communities, called \emph{islands} in our terminology. Each island has a set of agents implementing policies that should, if the training was successful, function well together.
    
    \item Each individual is a member of a \emph{species}. All individuals of the same species share a policy neural network.
    
    \item The algorithm unfolds on two timescales corresponding to (A) the population dynamics (ecological) time, and (B) policy execution (behavioral) time.
    
    \item The population dynamics are linked to individual reinforcement learning returns. If individuals of a given species perform well on a particular island then their population will increase there in the future.
    
    \item During each episode all individuals of a given species generate experience to train a common neural network via v-trace \cite{pmlr-v80-espeholt18a}. Experiences generated by individuals of a particular species are used only to update their own species neural network. After each episode the distribution over islands of each species is updated by a policy gradient-like update rule.

    \item \emph{Conservation of compute: } Biologically realistic population dynamics all contain at least the possibility of exponential growth. In practice, they are limited by carrying capacities, i.e., by environment properties. Since Malthusian reinforcement learning is mainly intended for multi-agent machine learning applications rather than for ecological simulations we cannot rely on resource constraints in the environment to limit growth. Thus, to ensure it can be executed with bounded compute resources, the population dynamic works by maintaining probability distributions for each species over the set of all islands. The probability assigned to any given island may grow or shrink based on the individual returns achieved there, but it is always constrained to be a valid probability distribution. Compute remains bounded because a fixed number of samples are used to assign individuals to islands. The total population varies on any given island from episode to episode, but across the entire \emph{archipelago}, the number of individuals is always constant.
\end{enumerate}

\begin{figure}[h]
\begin{tabular}{ l c r }
  \toprule
  
  {\bf Archipelago} & &\\
  $I$ & the set of islands in the archipelago &\\
  $N_I$ & the number of islands in an archipelago &\\
  $i$ & indexes islands &\\
  $e$ & indexes ecological scale time &\\
  
  \midrule
  {\bf Species} & &\\
  $L$ & the total number of species &\\
  $\Psi_l$ & a species &\\
  $\pi^l$ & the policy network of a species &\\
  $\theta^l$ & the parameters of $\pi^l$  &\\
  $\mu^l$ & the distribution of species $l$ over the archipelago &\\
  $w^l$ & the parameters of $\mu^l$ &\\
  $\Delta_{N_I}$ & the set of distributions over the archipelago &\\
  $K$ & total number of individuals &\\
  $M = \frac{K}{L}$ & number of species $l$ individuals across all islands &\\
  $k^l$ & labels individual $k$ of species $l$ &\\
  
  \midrule
  {\bf Population} & &\\
  $\Psi^l_{i,e}$ & the set of individuals of species $l$&\\
  & allocated to island $i$ at time-step $e$ &\\
  $\phi_e^{l,i}$ & the average fitness received by species $l$ &\\
  & on island $i$ at time $e$ &\\ 
  $\phi_{k^l,e}$ & the fitness received by individual $k^l$ of &\\
  & species $l$ at time $e$ &\\ 
  $\eta$ & population entropy regularization weight &\\
  $\alpha$ & population adaptation rate &\\
  
  \midrule
  {\bf Island} & &\\ 
  $t$ & indexes behavioral scale time & \\
  $N$ & the number of agents & \\
  $S$ & the state space &\\ 
  $s$ & a state &\\ 
  $A^i$ & the action space of player $i$ &\\ 
  $a^i$ & an action of player $i$ &\\
  $o^i$ & the observation of player $i$ &\\
  $\psi^i(\cdot)$ & the function that maps $s$ &\\
  & to the observation $o^i$ of player $i$ &\\
  $p(s_{t+1}|s_t,a^1,\dots,a^N)$ & is the transition kernel &\\
  $r^i(s,a^1,\dots,a^N)$ & the reward of player $i$ &\\
  \bottomrule
\end{tabular}
\end{figure}
\subsection{Archipelagos, islands, and species}
\paragraph{Island and Archipelago} an \textit{island} is a multi-agent environment where a variable number of agents can interact. An \textit{archipelago} is a set $I$ of islands. Furthermore we will write $N_I = |I|$ the number of islands in an archipelago.

\paragraph{Species} A species $\Psi^l$ is a set of individuals sharing the same policy network parameterization. There are $L$ species indexed by $l$. Each species is composed of a policy network $\pi^l$ with parameters $\theta^l$ which encodes the behavior of each individual of the species. The distribution of agents of a given species $l$ over the islands is $\mu^l \in \Delta_{N_I}$ (where $\Delta_{N_I}$ is the set of distributions over islands. $\mu^l$ is defined as a softmax over weights $w^l$. The total number of individuals is $K$ and the number of individuals per species is $M = \frac{K}{L}$. We will denote each individual of a species $l$ by $k^l$.

\begin{figure*}[t]
    \centering
    \includegraphics[width=\textwidth]{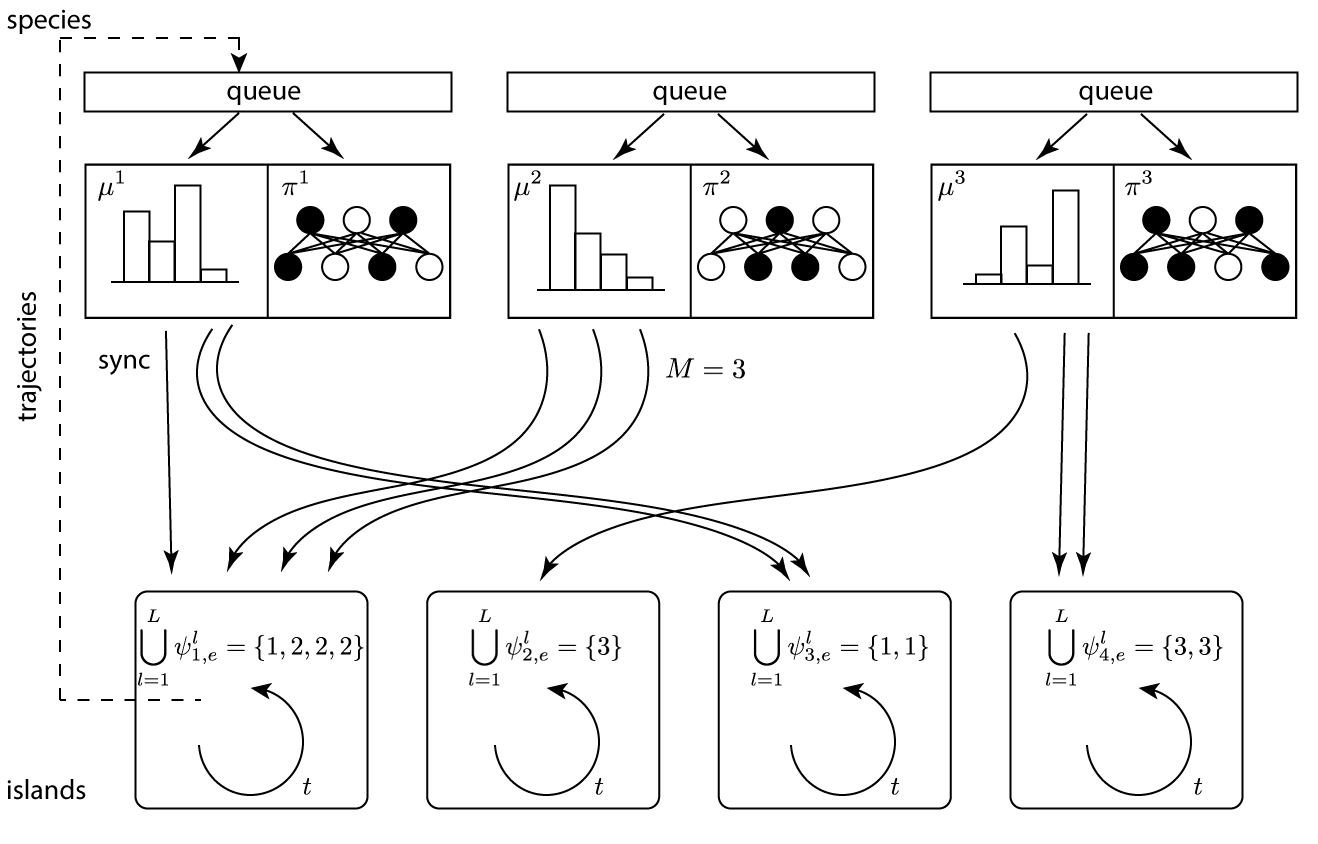}
    \caption{At each iteration on the ecological timescale $e$, each island samples the players to participate in its next episode according to the probability distributions $\mu^l$ over islands maintained by each species. Experienced trajectories from all conspecifics, on all islands, are used to update the same species policy network $\pi^l$. The distributions of returns to each species over islands are used to update the distributions from which to sample players for the next episode.}
    \label{fig:malthusian_illustration}
\end{figure*}

The learning process unfolds over two timescales, a slow \emph{ecological} scale which adapts the distribution of species over islands $\mu^l$ and a fast \emph{behavioral} scale over which individuals execute their policies. Species adapt $\pi^l$ to behave in the presence of others at the level of the island. The ecological scale timesteps are indexed by $e$, and the behavioral scale timesteps are indexed by $t$. The ecological scale ticks at the level of single episodes for the behavioral scale.

\subsection{Population dynamics}

The population dynamics govern how individuals of each species are assigned to the different islands over the ecological time scale. At a fixed ecological timestep $e$, individuals of each species are assigned to islands by sampling $M$ times from the distributions $\mu^l$. For each island, this yields an allocation $\Psi^l_{i,e}$, the set of individuals from species $l$ playing on island $i$ at ecological timestep $e$.

Each island has its own environment, in general the islands could have different environments from one another, though in this work we only consider the case where they are all the same.

Over the course of ecological time, the population evolves according to a gradient-based dynamic. At each timestep $e$, each individual $k^l$ of each species $l$ receives a fitness $\phi_{k^l, e}$, which is exactly its cumulative reward over the behavioral scale timesteps that have elapsed during one step of the ecological timescale. The per-island fitness for each species is then calculated as

$$\phi_{i, e}^l = \left(\sum \limits_{k^l \in \Psi^l_{i, e}} \phi_{k^l, e} \right) / |\Psi^l_{i, e}| \textrm{  and  } 0 \textrm{  if  } \Psi^l_{i, e} = \emptyset \, .$$

The distribution over islands for each species, $\mu^l(i) = e^{w^l_i} / \sum_j e^{w^l_j}$, is updated according to policy gradient with entropy regularization. Explicitly the distribution weights for species $l$ over all islands change according to a policy-gradient update
$$w^l_{e+1} = w^l_e + \alpha \left[  \sum \limits_{i \in \{1, \dots, N_I\}} \nabla_{w^l} \mu^l(i) ( \phi_{i,e}^l  - \eta  \log\mu^l(i))\right] \,.$$

The goal of the entropy regularization term is to enforce that some minimal population of each species remains on sub-optimal islands. Thus, the population distributions adapt over ecological time so as to minimize the following loss:

$$\left[  \sum \limits_{i \in \{1, \dots, N_I\}} \mu^l(i) ( \phi_{i,e}^l  - \eta  \log\mu^l(i))\right].$$

\subsection{Multiagent Reinforcement Learning}

A Partially Observable Markov Game (POMG) is sequential decision model of a multiagent environment in which $N$ individuals interact. At each state $s \in S$ of a POMG, each agent selects an action $a^i \in A^i$ based on the observation $o^i$ of the state of the game they have. The observation of player $i$ is defined here as a function of the state $o^i = \psi^i(s)$. Then the state changes to $s' \sim p(.|s,a^1,\dots,a^N)$ and the individuals receive reward $r^i(s,a^1,\dots,a^N)$. Each species learns a policy $\pi^l(a^i|o^i)$ given the experience of each of its individuals. At each step, all individuals of a species collect trajectories of the experience gathered in the island they have been assigned to. The reinforcement learning algorithm produces gradient updates of the parameters for each individual of the species. The gradient updates are then averaged over all individuals of the species to update the parameters $\{\theta_l\}_{\{1,\dots,L\}}$. The V-trace algorithm is used to update the parameters as described in \cite{pmlr-v80-espeholt18a} with truncation levels set to $1$. Note that experience (observations, actions and rewards) from all individuals of a species contribute equally, but that the individuals may be spread non-uniformly over islands. This means that the species parameter update may be disproportionately affected by the performance of the species on particular islands.

\begin{figure}[h]
\begin{tabular}{ l c r }
  \toprule
  {\bf RL Agent} & &\\
    LSTM Unroll Length:& $20$&\\
    Entropy Regularizer:& $\sim \text{log-uniform}(0.00005, 0.05)$&\\
    Baseline loss scaling:& $0.5$&\\
    Discount:& $0.99$&\\
  {\bf Optimization} & &\\
    RMSProp learning rate: & $\sim \text{log-uniform}(0.0001, 0.005)$ &\\
    RMSProp $\epsilon$: & $0.0001$ &\\
    RMSProp decay: & $0.99$ &\\
    Batch size: & $32$ &\\
  \bottomrule
\end{tabular}
\end{figure}

\paragraph{Function approximation:}
The neural network architecture was similar to that of \cite{mnih2016asynchronous}. It consists of a convnet with $16$ channels, kernel size of $3$, and stride of $1$. The output of the convnet is passed to a a 1-layer MLP of size $32$, followed by a recurrent module (an LSTM \cite{Hochreiter:1997:LSM:1246443.1246450}) of size $64$. The recurrent module's output is then linearly transformed into the the policy and value. All nonlinearities between layers were rectified linear units.

\paragraph{Distributed computing:}
The island simulation and the species neural network updates were implemented as separate processes, potentially running on different machines. Islands produce trajectories and send them to a circular queue on the species update process. The species update process waits until it can dequeue a complete  batch of $32$ trajectories, at which point it computes the v-trace update.

\paragraph{Environments:}
The games studied in this work are all partially observable in that individuals can only observe via a 15$\times$15 RGB window, centered on their current location. The action space consists of moving left, right, up, and down, rotating left and right. Each species was assigned a unique color, shared by all conspecifics and preserved across all islands.

\section{Results}

\subsection{Exploration experiments}

Given an unrefined and infrequently emitted behavior, reinforcement learning algorithms are very good at estimating its value with respect to alternatives and refining it into a well-honed strategy for achieving rewards. However, a central problem in reinforcement learning concerns the initial origin of such behaviors, especially in cases where the state space is too large for exhaustive search, and there are many local optima where the policy's reward gradient becomes zero.

\begin{figure*}[t]
    \centering
    \includegraphics[width=\textwidth]{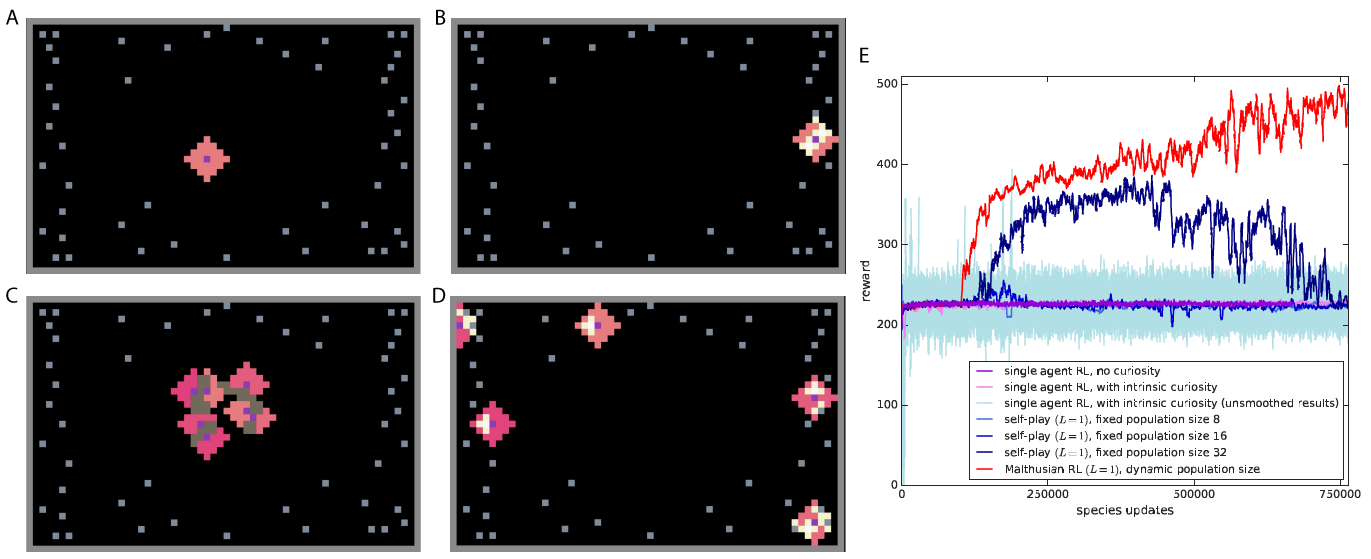}
    \caption{Experiments with extrinsically and intrinsically motivated individual exploration using the Clamity game. (A) Local optimum outcome. (B) Global optimum outcome. (C) Catastrophic multi-player outcome. (D) Multi-player global optimum. (E) Returns as a function of the number of times the species playing on the evaluated solitary island was updated. Except where indicated otherwise. reward values were smoothed over time with a window size of 100. Malthusian RL parameters were $\alpha = 0.0001$ and $\eta = 1.5$. Each episode lasted $250$ behavior steps.}
    \label{fig:clamity}
\end{figure*}

This section explores how population dynamics may drive innovation in individual behavior. To study this, we introduce a new game that taxes individual exploration skills. It can be seen as a multi-agent analog of the well-known Montezuma's Revenge single player game that has often been used for studies of intrinsic motivations for single-agent exploration \cite{bellemare2013arcade, bellemare2016unifying, ostrovski2017count, martin2017count, conti2017improving}. We hold to the game theory tradition of introducing each game with a facetious (but hopefully memorable) story, and offer the following:

In the \emph{Clamity} game, agents begin in the trochophore stage of the bivalve mollusk lifecycle. They can freely swim around the map, a partially observed grid-world (map size = $36 \times 60$, window size = $15 \times 15$. Then whenever they are ready, they can perform the *settle* action. This action causes the agent to metamorphose into the adult clam stage of their lifecycle at their current location and removes their ability to swim. After settling, their shell grows around them, up to a maximum size. Shell growth is also restricted by the presence of adjacent shells from other clams. Each adult clam filters invisible food particles from the ocean at a rate proportional to the size of its shell, receiving reward for each food particle filtered. However, clam shells that are adjacent to the shell of another individual become unhealthy and do not filter any food. There are also nutrient patches located a considerable distance away from the starting location (more than $10$ steps away, see maps in Fig. \ref{fig:clamity}-A). Individuals that settle near a nutrient patch so that it is either partially or fully engulfed in their shell absorb additional nutrients from it. Episodes terminate after $T = 250$ steps. Settling immediately on the first action is a very attractive local optimum. The global optimum solution is to swim quickly out to a nutrient patch and settle there instead\footnote{A video of the single-agent global optimum policy can be viewed here: \newline { \url{https://youtu.be/AIT3FTC9s4s}}.}.

Single-agent reinforcement learning algorithms become stuck in the local optimum\footnote{A video of the single-agent local optimum policy can be viewed here: \newline { \url{https://youtu.be/OHkpe9dVGyw}}.} and fail to ever discover the nutrient patches. To see why, consider the number of consecutive seemingly suboptimal actions that an agent would have to take in order to discover a nutrient patch. The settle action can be taken at any time, it always provides some level of rewards, and once taken, prevents movement for the rest of the episode. Thus any reasonable reinforcement learning algorithm that follows the initial gradient of its experience will reach the local optimum. If it starts out settling on step $t_s > 1$, it will receive an expected return of $T - t_s \times \text{reward rate}$. But if it were to settle earlier instead, e.g., on step $t_s - 1$, it would receive a larger expected return. Thus there is a strong gradient from any policy initialization to the local optimum of settling on step $t_s = 1$. Furthermore. since the environment is partially observable, a single agent would need to choose to move in the same direction for several  steps despite registering no change at all in its observation during that time.

On the other hand, Clamity can also be played by multiple agents simultaneously. All the trochophores begin each episode nearby one another in the center of the map. Since intersecting shells become unhealthy and provide no reward, individuals are penalized for settling too close to one another\footnote{A video of such a multiplayer bad outcome can be viewed here: \newline {\url{https://youtu.be/vrXOtHYMaPE}}.}. This provides a gradient that incentivizes agents to swim away from the starting location to avoid competing with one another for shell space. If the population size is large enough then this competition-motivated spreading eventually leads individuals to discover the nutrient patches\footnote{A video of a group of agents implementing a multiplayer global optimum joint policy can be viewed here:  {\url{https://youtu.be/TnxMnSClBHY}}.}.

\subsubsection{Experimental procedure}

To make like-for-like comparisons between single-agent and multi-agent training regimes, we adopt the following protocol. In parallel with the archipelago ($N_I$ islands), we run $L$ (the number of species) additional \emph{solitary islands}. On the $l$-th solitary island, a single individual of species $l$ plays each episode alone. All the experience generated on islands where species $\Psi_l$ appears, even its solitary island, is used to update its policy $\pi^l$. However, the amount of each species's total experience derived from the solitary island is comparatively small since in this experiment, $M = 960$, the number of individuals of species $l$ appearing across all islands of the archipelago. The final results are reported only from the solitary islands but reflect the policy learning accumulated in the competitive archipelago setting.

Our single agent training protocol simply sets the number of islands in the archipelago $N_I$ to $0$ and replicates each solitary island $32$ times. Since there is only a single species ($L = 1$), and all solitary island replicas are the same as one another (though with different random environment seeds), the result is exactly equivalent to the A3C training regime \cite{mnih2016asynchronous}.

This protocol also makes it easy to compare the proposed training regime where population sizes are dynamic and variable from episode to episode to the case of a ``standard self-play'' training regime, where population sizes are fixed. In this case, the archipelago contains just one island inhabited by a fixed number of individuals. As before, most of the experience is generated from the island where multiple individuals play. Results are reported only from the (single) solitary island, just as it is in the dynamic case.

\subsubsection{Results}

Individuals of species trained by Malthusian reinforcement learning find the globally optimum single-player solution, despite most of their experience coming from multi-player islands. Individuals trained by two baseline single-agent reinforcement learning algorithms completely fail to escape the local optimum. The first baseline we tried had all the exact same hyperparameters as in the Malthusian case, but all of its experience was in solitary islands ($32$ of them in parallel).\footnote{These hyperparameters were not tuned for the Malthusian case---they were prespecified before the runs of both methods, and not subsequently changed.} The second single-agent reinforcement learning algorithm baseline we tried was an implementation of the current state-of-the-art in curiosity-driven reinforcement learning, the intrinsic curiosity module provides a pseudo-reward to the agents based on its prediction error in predicting the next timestep in the evolution of a compressed encoding of its observations \cite{pathak2017curiosity, martin2017count}. In this case, augmenting the agent with the intrinsic curiosity module is still insufficient to get it it to consistently discover the nutrient patches. It does stumble upon them from time to time, especially early on in training (Fig.~\ref{fig:clamity}, but does not even do so consistently enough to register in a smoothed plot of rewards versus time with a $100$ step smoothing window (Fig.~\ref{fig:clamity}). In contrast, individual members of species trained by Malthusian reinforcement learning with dynamic population sizes consistently implement globally optimal policies once they have discovered them (Fig.~\ref{fig:clamity}). 

Next we asked whether dynamic population sizes were specifically important or whether the key was just the simultaneous training in multi-agent islands with a given, sufficiently large, population size. We noticed that most runs with dynamical population sizes converged to an island population size around $32$ in the best performing islands. Thus we ran several experiments where agents trained in fixed population islands, evaluated on solitary islands as before. We found that individuals that trained in a fixed population size of $32$ were able to discover the global optimum, but apparently less consistently than in the case with dynamic population size (red curve above navy curve in Fig.~\ref{fig:clamity}), and apparently with greater vulnerability to forgetting (the navy colored curve eventually declines back to the local optimum).

\begin{figure*}[t]
    \centering
    \includegraphics[width=\textwidth]{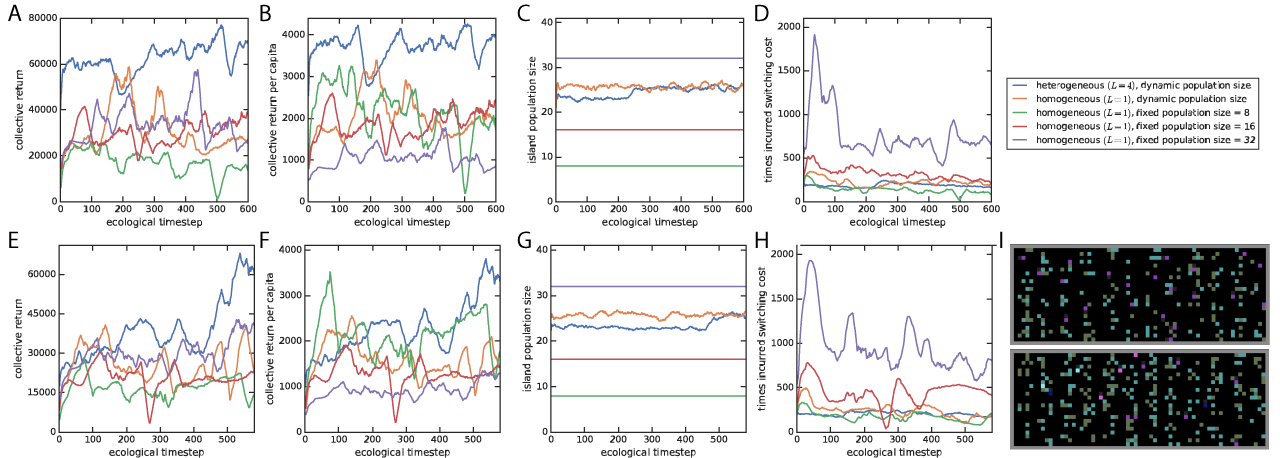}
    \caption{Experiments with the evolution of mutualism using the Allelopathy game. All results in this figure were smoothed with a window size of 25 ecological steps. (A-D) Unbiased Allelopathy game. Malthusian RL parameters were $\alpha = 1e-07$ and $\eta = 0.3$. E-H) Biased Allelopathy game. Malthusian RL parameters were $\alpha = 0.0001$ and $\eta = 0.01$. (A, E) Maximum collective return over all islands as a function of ecological time. (B, F) Maximum per capita collective return over all islands as a function of ecological time. (C, G) Maximum island population size over all islands as a function of ecological time. (D, H) Minimum number of times incurred a switching cost as a function of ecological time. ((I) Two screenshots of random procedurally generated initial map configurations. Maps were procedurally generated by randomly placing shrubs at the start of each episode. Episodes lasted $1000$ behavior steps.}
    \label{fig:allelopathy_results}
\end{figure*}

\subsection{Mutualism and specialization experiments}

Solution concepts for general-sum games may involve mutualistic interactions between synergistic strategies. Successful cooperative joint strategies may be either homogeneous, as in facultative mutualism, or heterogeneous. In nature, partners in mutually profitable associations are often very different from one another so that they can provide complementary capabilities to the partnership. In fact, most known mutualisms involve partners from different kingdoms, e.g., corals and their algae symbionts, vascular plants and mycorrhizal fungi, mammals and their gut bacteria, etc \cite{bruno2003inclusion}. Moreover, division of labor and the subsequent efficiency gains from specialization are thought to be key components of complex human society \cite{smith1776inquiry}. 

However, it may be difficult to learn such mutually profitable partnerships of widely divergent strategies with self-play. All partners would need to represent all specializations, wasting valuable representation capacity. In addition, a policy learned by self-play requires a switching mechanism to break the symmetry and determine which sub-policy to emit in any given situation. For example, an agent could learn to become a blacksmith if standing on the left and a farmer if standing on the right. The complexity of the switching policy is itself related to the extent of partial observability in the environment. In some cases it may be very difficult to determine the right proportion of individuals needed to perform each part of the partnership at any given time, e.g. if the others' strategies cannot easily be observed. It would be easier to learn a heterogeneous set of policies, each one implementing only its own part of the partnership. But then, it would seem that the number of copies of each would have to be known in advance, thus adding many new difficult-to-tune hyperparameters, one for each species. 

In this section we explore whether Malthusian reinforcement learning can find mutualistic partnerships more easily than other multi-agent reinforcement learning methods, especially when there is a potential for gains from heterogeneous populations containing multiple specialized members. To test this, we created another partially observed Markov game environment. Again continuing the game-theoretic tradition of accompanying each game with a facetious and memorable story, we offer the following.

The Allelopathy game has two main rules. (1) shrubs grow in random positions on an open field. Shrubs allelopathically suppress one another's growth. That is, the probability that a seed of a given type grows into a shrub in any given timestep is inversely proportional to the number of nearby shrubs of other types. (2) Agents in Allelopathy are herbivorous animals that can eat many different types of shrub. However, switching frequently between digesting different shrub types imposes a metabolic cost since different enzymes must be synthesized for each. Thus, agents benefit from specialization in eating only a single type of shrub. Agents receive increasing rewards for repeatedly harvesting the same type of shrub (up to a maximum of $r = 250$). Rewards drop back down to their lowest level, $(r = 1)$, when the agent harvests a different type of shrub (since that entails their switching into a different metabolic regime). Thus an agent that randomly harvests any shrub they come across is likely to receive low rewards. An agent that only harvests a particular type of shrub while ignoring others will obtain significantly greater rewards. The combined effect of these two rules is to make it so that a specialist in any one shrub type benefits from the presence of others who specialize in different shrub types since their foraging increases the growth rate of all the shrubs they do not consume.

We studied two variants of the Allelopathy game. The first variant, unbiased Allelopathy, has two shrub types $A$ and $B$ that appear with equal probability. In the second variant, biased Allelopathy, the two shrub types do not appear with the same frequency. Type $A$ is significantly more common than type $B$. In addition, each shrub of type $A$ consumed provides a maximum reward of $8$ when at least $8$ in a row are consumed.  Whereas type $B$ shrubs yield a maximum reward of $250$ for any agent that manages to consume that many consecutively. Biased Allelopathy is a social dilemma since specialists in type $B$ are clearly better off than specialists in type $A$, but both do better when the other is around.

\subsubsection{Results}

\begin{figure}[h]
    \centering
    \includegraphics[width=\columnwidth]{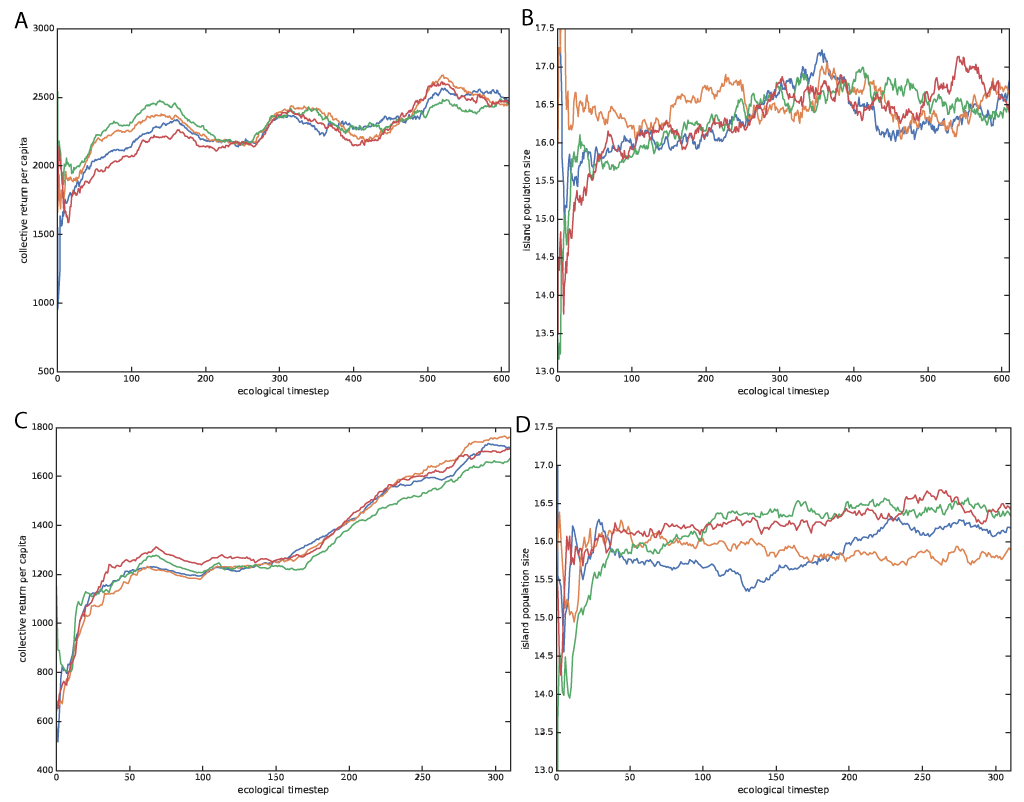}
    \caption{Representative island timecourses for the Allelopathy game. The different lines represent different islands. Notice that the results are consistent across islands. (A-B) results from the unbiased Allelopathy game. (C-D) Results from the biased Allelopathy game. (A, C) Collective return per capita as a function of ecological time for four representative islands. (B, D) Island population size as a function of ecological time for four representative islands.}
    \label{fig:allelopathy_secondary}
\end{figure}

Here the critical comparison is between homogeneous $(L = 1)$ and heterogeneous $(L > 1)$ population dynamics. Therefore the object of study is the performance of the \textit{islands} rather than specific individuals. The Allelopathy environment contains two niches, corresponding to specialization in consuming either shrub type $A$ or $B$. In the heterogeneous case, mutualistic partnerships may develop from initial conditions where species in proximity to one another randomly fill either role. This situation features a gradient that guides each species in different directions. Whichever species begins with a propensity toward role $A$ ends up specializing in role $A$. Likewise, the other species evolves to specialize in role $B$, to the mutual benefit of both partners. On the other hand, in the homogeneous case, it is still possible for mutualistic interactions to develop, but it is more difficulty since (1) both specialized parts of the joint policy must be represented in the same network, and (2) the policy must include a switching mechanism that breaks the symmetry, determining which sub-policy to implement in each situation. Heterogeneous species avoid the need for this symmetry breaking, and the relative proportions assigned to each role are handled naturally by the adapting relative population sizes (Fig.~\ref{fig:allelopathy_secondary}).

The total number of individuals in each experiment was $K = 960$. Thus in the homogeneous $L = 1$ case, $M = 960$, and in the heterogeneous $L = 4$ case, $M = 240$. The number of islands $N_I$ was $60$ in both dynamic population size conditions. In the fixed population size conditions the number of islands was chosen so that the total number of individual instances would still be $960$, e.g., for fixed population size $32$, this required $N_I = 960 / 32 = 30$.

Results were similar for both the biased and unbiased Allelopathy games. Heterogeneous $(L = 4)$ population dynamics achieved higher returns, both per capita, and in aggregate, than the other tested methods including the homogeneous population $(L = 1)$ with size dynamics (Fig.~\ref{fig:allelopathy_results}). Interestingly both heterogeneous and homogeneous runs converged to the same population size, but the heterogeneous case increased more slowly to that point, and did so while maintaining a higher per capita rate of return.

\section{Discussion}

This paper introduces Malthusian reinforcement learning, a multi-agent reinforcement learning algorithm that motivates individual exploration and takes advantage of possibilities for synergy to evolve heterogeneous mutualisms. If populations rise when returns improve then the problem itself shifts over time so no local optimum need ever be reached. This gives rise to a strategy that we may term \emph{exploration by exploitation}. Individuals can always follow the gradient of their experience, they need never depart from their current estimate of the best policy just to explore the state space. They will naturally explore it, just by following the gradient in a changing world. Opportunities for heterogeneous mutualism may also be detected by gradient following. Initially weak specialization in one agent incentivizes its soon-to-be partner to specialize in a complementary direction, which in turn catalyzes more specialization, and so on.

How does this paper's proposed population dynamic relate to dynamics studied in evolutionary theory? Our requirement of conservation of compute, that the number of individuals of a given species on a given island may vary from episode to episode, but the total number of individuals of each species in the  archipelago is always the same fixed value, implies that for populations to increase on one island they much decrease elsewhere. Thus the population dynamic introduced here may be understood as an evolutionary model of migration. Moreover, since fitnesses are computed globally, i.e. relative to the entire archipelago, it is more similar to \emph{hard} selection models in evolutionary theory where populations are regulated globally than \emph{soft} selection, where population regulation occurs locally within each island \cite{doi:10.1086/284328, West72, henrich2004}.

Other possible relationships between population size and innovation have appeared in the evolutionary anthropology literature. For instance, it is possible that---especially in preliterate societies---larger populations provide for more protection from forgetting of useful cultural elements since more elders, functioning as repositories of cultural knowledge, will be alive at any given time \cite{henrich2004demography}. Or alternatively, larger social networks may provide more opportunities for recombination of disparate cultural elements that originated in farther and farther away contexts \cite{boyd2011cultural, kempe2014cultural, henrich2010markets, muthukrishna2016innovation}. As hypotheses for the origin of innovative behaviors in biology, these possibilities appear to be at odds with the mechanism implied by our algorithm, since each could explain, for instance, the same correlations between brain size and group size across the primate order \cite{dunbar2017there} as well as innovation and (cultural) group size in humans \cite{richerson2009cultural, muthukrishna2016innovation}. However, they are not mutually exclusive. In fact, all three mechanisms may even operate synergistically with one another. More research is needed in order to tease apart the precise mechanisms in biology. In computer science, we think this line of thought opens up a goldmine of new algorithmic ideas concerning the combination of population dynamics with social learning and imitation.

%%%%%%%%%%%%%%%%%%%%%%%%%%%%%%%%%%%%%%%%%%%%%%%%%%%%%%%%%%%%%%%%%%%%%%%%%%%%%%%%%%%%%%%%%%%%%%%%%%%%%%%%%
%% Acknowledgements are not in the review version.
\begin{acks}
  We would like to thank Tina Zhu for coming up with the name ``Clamity'' and Oliver Smith for program management support.
\end{acks}
%%%%%%%%%%%%%%%%%%%%%%%%%%%%%%%%%%%%%%%%%%%%%%%%%%%%%%%%%%%%%%%%%%%%%%%%%%%%%%%%%%%%%%%%%%%%%%%%%%%%%%%%%
%% bibliography: see CFP for number of permitted pages

\bibliographystyle{ACM-Reference-Format}  % do not change this line!
\balance  % do not change this line -- unless you manually balance your last page
\bibliography{main}  % put name of your .bib file here

%%% -*-BibTeX-*-
%%% Do NOT edit. File created by BibTeX with style
%%% ACM-Reference-Format-Journals [18-Jan-2012].

\begin{thebibliography}{00}

%%% ====================================================================
%%% NOTE TO THE USER: you can override these defaults by providing
%%% customized versions of any of these macros before the \bibliography
%%% command.  Each of them MUST provide its own final punctuation,
%%% except for \shownote{}, \showDOI{}, and \showURL{}.  The latter two
%%% do not use final punctuation, in order to avoid confusing it with
%%% the Web address.
%%%
%%% To suppress output of a particular field, define its macro to expand
%%% to an empty string, or better, \unskip, like this:
%%%
%%% \newcommand{\showDOI}[1]{\unskip}   % LaTeX syntax
%%%
%%% \def \showDOI #1{\unskip}           % plain TeX syntax
%%%
%%% ====================================================================

\ifx \showCODEN    \undefined \def \showCODEN     #1{\unskip}     \fi
\ifx \showDOI      \undefined \def \showDOI       #1{#1}\fi
\ifx \showISBNx    \undefined \def \showISBNx     #1{\unskip}     \fi
\ifx \showISBNxiii \undefined \def \showISBNxiii  #1{\unskip}     \fi
\ifx \showISSN     \undefined \def \showISSN      #1{\unskip}     \fi
\ifx \showLCCN     \undefined \def \showLCCN      #1{\unskip}     \fi
\ifx \shownote     \undefined \def \shownote      #1{#1}          \fi
\ifx \showarticletitle \undefined \def \showarticletitle #1{#1}   \fi
\ifx \showURL      \undefined \def \showURL       {\relax}        \fi
% The following commands are used for tagged output and should be
% invisible to TeX
\providecommand\bibfield[2]{#2}
\providecommand\bibinfo[2]{#2}
\providecommand\natexlab[1]{#1}
\providecommand\showeprint[2][]{arXiv:#2}

\bibitem[\protect\citeauthoryear{Bansal, Pachocki, Sidor, Sutskever, and
  Mordatch}{Bansal et~al\mbox{.}}{2017}]%
        {bansal2017emergent}
\bibfield{author}{\bibinfo{person}{Trapit Bansal}, \bibinfo{person}{Jakub
  Pachocki}, \bibinfo{person}{Szymon Sidor}, \bibinfo{person}{Ilya Sutskever},
  {and} \bibinfo{person}{Igor Mordatch}.} \bibinfo{year}{2017}\natexlab{}.
\newblock \showarticletitle{Emergent complexity via multi-agent competition}.
\newblock \bibinfo{journal}{{\em arXiv preprint arXiv:1710.03748\/}}
  (\bibinfo{year}{2017}).
\newblock


\bibitem[\protect\citeauthoryear{Bellemare, Srinivasan, Ostrovski, Schaul,
  Saxton, and Munos}{Bellemare et~al\mbox{.}}{2016}]%
        {bellemare2016unifying}
\bibfield{author}{\bibinfo{person}{Marc Bellemare}, \bibinfo{person}{Sriram
  Srinivasan}, \bibinfo{person}{Georg Ostrovski}, \bibinfo{person}{Tom Schaul},
  \bibinfo{person}{David Saxton}, {and} \bibinfo{person}{Remi Munos}.}
  \bibinfo{year}{2016}\natexlab{}.
\newblock \showarticletitle{Unifying count-based exploration and intrinsic
  motivation}. In \bibinfo{booktitle}{{\em Advances in Neural Information
  Processing Systems}}. \bibinfo{pages}{1471--1479}.
\newblock


\bibitem[\protect\citeauthoryear{Bellemare, Naddaf, Veness, and
  Bowling}{Bellemare et~al\mbox{.}}{2013}]%
        {bellemare2013arcade}
\bibfield{author}{\bibinfo{person}{Marc~G Bellemare}, \bibinfo{person}{Yavar
  Naddaf}, \bibinfo{person}{Joel Veness}, {and} \bibinfo{person}{Michael
  Bowling}.} \bibinfo{year}{2013}\natexlab{}.
\newblock \showarticletitle{The arcade learning environment: An evaluation
  platform for general agents}.
\newblock \bibinfo{journal}{{\em Journal of Artificial Intelligence
  Research\/}}  \bibinfo{volume}{47} (\bibinfo{year}{2013}),
  \bibinfo{pages}{253--279}.
\newblock


\bibitem[\protect\citeauthoryear{Boyd, Richerson, and Henrich}{Boyd
  et~al\mbox{.}}{2011}]%
        {boyd2011cultural}
\bibfield{author}{\bibinfo{person}{Robert Boyd}, \bibinfo{person}{Peter~J
  Richerson}, {and} \bibinfo{person}{Joseph Henrich}.}
  \bibinfo{year}{2011}\natexlab{}.
\newblock \showarticletitle{The cultural niche: Why social learning is
  essential for human adaptation}.
\newblock \bibinfo{journal}{{\em Proceedings of the National Academy of
  Sciences\/}} \bibinfo{volume}{108}, \bibinfo{number}{Supplement 2}
  (\bibinfo{year}{2011}), \bibinfo{pages}{10918--10925}.
\newblock


\bibitem[\protect\citeauthoryear{Bruno, Stachowicz, and Bertness}{Bruno
  et~al\mbox{.}}{2003}]%
        {bruno2003inclusion}
\bibfield{author}{\bibinfo{person}{John~F Bruno}, \bibinfo{person}{John~J
  Stachowicz}, {and} \bibinfo{person}{Mark~D Bertness}.}
  \bibinfo{year}{2003}\natexlab{}.
\newblock \showarticletitle{Inclusion of facilitation into ecological theory}.
\newblock \bibinfo{journal}{{\em Trends in Ecology \& Evolution\/}}
  \bibinfo{volume}{18}, \bibinfo{number}{3} (\bibinfo{year}{2003}),
  \bibinfo{pages}{119--125}.
\newblock


\bibitem[\protect\citeauthoryear{Burda, Edwards, Pathak, Storkey, Darrell, and
  Efros}{Burda et~al\mbox{.}}{2018}]%
        {burda2018large}
\bibfield{author}{\bibinfo{person}{Yuri Burda}, \bibinfo{person}{Harri
  Edwards}, \bibinfo{person}{Deepak Pathak}, \bibinfo{person}{Amos Storkey},
  \bibinfo{person}{Trevor Darrell}, {and} \bibinfo{person}{Alexei~A Efros}.}
  \bibinfo{year}{2018}\natexlab{}.
\newblock \showarticletitle{Large-scale study of curiosity-driven learning}.
\newblock \bibinfo{journal}{{\em arXiv preprint arXiv:1808.04355\/}}
  (\bibinfo{year}{2018}).
\newblock


\bibitem[\protect\citeauthoryear{Chentanez, Barto, and Singh}{Chentanez
  et~al\mbox{.}}{2005}]%
        {chentanez2005intrinsically}
\bibfield{author}{\bibinfo{person}{Nuttapong Chentanez},
  \bibinfo{person}{Andrew~G Barto}, {and} \bibinfo{person}{Satinder~P Singh}.}
  \bibinfo{year}{2005}\natexlab{}.
\newblock \showarticletitle{Intrinsically motivated reinforcement learning}. In
  \bibinfo{booktitle}{{\em Advances in neural information processing systems}}.
  \bibinfo{pages}{1281--1288}.
\newblock


\bibitem[\protect\citeauthoryear{Clark}{Clark}{2008}]%
        {clark2008farewell}
\bibfield{author}{\bibinfo{person}{Gregory Clark}.}
  \bibinfo{year}{2008}\natexlab{}.
\newblock \bibinfo{booktitle}{{\em A farewell to alms: a brief economic history
  of the world}}. Vol.~\bibinfo{volume}{27}.
\newblock \bibinfo{publisher}{Princeton University Press}.
\newblock


\bibitem[\protect\citeauthoryear{Conti, Madhavan, Such, Lehman, Stanley, and
  Clune}{Conti et~al\mbox{.}}{2017}]%
        {conti2017improving}
\bibfield{author}{\bibinfo{person}{Edoardo Conti}, \bibinfo{person}{Vashisht
  Madhavan}, \bibinfo{person}{Felipe~Petroski Such}, \bibinfo{person}{Joel
  Lehman}, \bibinfo{person}{Kenneth~O Stanley}, {and} \bibinfo{person}{Jeff
  Clune}.} \bibinfo{year}{2017}\natexlab{}.
\newblock \showarticletitle{Improving exploration in evolution strategies for
  deep reinforcement learning via a population of novelty-seeking agents}.
\newblock \bibinfo{journal}{{\em arXiv preprint arXiv:1712.06560\/}}
  (\bibinfo{year}{2017}).
\newblock


\bibitem[\protect\citeauthoryear{Dunbar and Shultz}{Dunbar and Shultz}{2017}]%
        {dunbar2017there}
\bibfield{author}{\bibinfo{person}{RIM Dunbar} {and} \bibinfo{person}{Susanne
  Shultz}.} \bibinfo{year}{2017}\natexlab{}.
\newblock \showarticletitle{Why are there so many explanations for primate
  brain evolution?}
\newblock \bibinfo{journal}{{\em Phil. Trans. R. Soc. B\/}}
  \bibinfo{volume}{372}, \bibinfo{number}{1727} (\bibinfo{year}{2017}),
  \bibinfo{pages}{20160244}.
\newblock


\bibitem[\protect\citeauthoryear{Eriksson, Betti, Friend, Lycett, Singarayer,
  von Cramon-Taubadel, Valdes, Balloux, and Manica}{Eriksson
  et~al\mbox{.}}{2012}]%
        {eriksson2012late}
\bibfield{author}{\bibinfo{person}{Anders Eriksson}, \bibinfo{person}{Lia
  Betti}, \bibinfo{person}{Andrew~D Friend}, \bibinfo{person}{Stephen~J
  Lycett}, \bibinfo{person}{Joy~S Singarayer}, \bibinfo{person}{Noreen von
  Cramon-Taubadel}, \bibinfo{person}{Paul~J Valdes}, \bibinfo{person}{Francois
  Balloux}, {and} \bibinfo{person}{Andrea Manica}.}
  \bibinfo{year}{2012}\natexlab{}.
\newblock \showarticletitle{Late Pleistocene climate change and the global
  expansion of anatomically modern humans}.
\newblock \bibinfo{journal}{{\em Proceedings of the National Academy of
  Sciences\/}} \bibinfo{volume}{109}, \bibinfo{number}{40}
  (\bibinfo{year}{2012}), \bibinfo{pages}{16089--16094}.
\newblock


\bibitem[\protect\citeauthoryear{Espeholt, Soyer, Munos, Simonyan, Mnih, Ward,
  Doron, Firoiu, Harley, Dunning, Legg, and Kavukcuoglu}{Espeholt
  et~al\mbox{.}}{2018}]%
        {pmlr-v80-espeholt18a}
\bibfield{author}{\bibinfo{person}{Lasse Espeholt}, \bibinfo{person}{Hubert
  Soyer}, \bibinfo{person}{Remi Munos}, \bibinfo{person}{Karen Simonyan},
  \bibinfo{person}{Vlad Mnih}, \bibinfo{person}{Tom Ward},
  \bibinfo{person}{Yotam Doron}, \bibinfo{person}{Vlad Firoiu},
  \bibinfo{person}{Tim Harley}, \bibinfo{person}{Iain Dunning},
  \bibinfo{person}{Shane Legg}, {and} \bibinfo{person}{Koray Kavukcuoglu}.}
  \bibinfo{year}{2018}\natexlab{}.
\newblock \showarticletitle{{IMPALA}: Scalable Distributed Deep-{RL} with
  Importance Weighted Actor-Learner Architectures}. In \bibinfo{booktitle}{{\em
  Proceedings of the 35th International Conference on Machine Learning}} {\em
  (\bibinfo{series}{Proceedings of Machine Learning Research})},
  \bibfield{editor}{\bibinfo{person}{Jennifer Dy} {and}
  \bibinfo{person}{Andreas Krause}} (Eds.), Vol.~\bibinfo{volume}{80}.
  \bibinfo{publisher}{PMLR}, \bibinfo{address}{Stockholmsmässan, Stockholm
  Sweden}, \bibinfo{pages}{1407--1416}.
\newblock


\bibitem[\protect\citeauthoryear{Goebel, Waters, and O'rourke}{Goebel
  et~al\mbox{.}}{2008}]%
        {goebel2008late}
\bibfield{author}{\bibinfo{person}{Ted Goebel}, \bibinfo{person}{Michael~R
  Waters}, {and} \bibinfo{person}{Dennis~H O'rourke}.}
  \bibinfo{year}{2008}\natexlab{}.
\newblock \showarticletitle{The late Pleistocene dispersal of modern humans in
  the Americas}.
\newblock \bibinfo{journal}{{\em science\/}} \bibinfo{volume}{319},
  \bibinfo{number}{5869} (\bibinfo{year}{2008}), \bibinfo{pages}{1497--1502}.
\newblock


\bibitem[\protect\citeauthoryear{Henrich}{Henrich}{2004a}]%
        {henrich2004}
\bibfield{author}{\bibinfo{person}{Joseph Henrich}.}
  \bibinfo{year}{2004}\natexlab{a}.
\newblock \showarticletitle{Cultural group selection, coevolutionary processes
  and large-scale cooperation}.
\newblock   \bibinfo{volume}{53} (\bibinfo{date}{02} \bibinfo{year}{2004}),
  \bibinfo{pages}{3--35}.
\newblock


\bibitem[\protect\citeauthoryear{Henrich}{Henrich}{2004b}]%
        {henrich2004demography}
\bibfield{author}{\bibinfo{person}{Joseph Henrich}.}
  \bibinfo{year}{2004}\natexlab{b}.
\newblock \showarticletitle{Demography and cultural evolution: how adaptive
  cultural processes can produce maladaptive losses---the Tasmanian case}.
\newblock \bibinfo{journal}{{\em American Antiquity\/}} \bibinfo{volume}{69},
  \bibinfo{number}{2} (\bibinfo{year}{2004}), \bibinfo{pages}{197--214}.
\newblock


\bibitem[\protect\citeauthoryear{Henrich, Ensminger, McElreath, Barr, Barrett,
  Bolyanatz, Cardenas, Gurven, Gwako, Henrich, et~al\mbox{.}}{Henrich
  et~al\mbox{.}}{2010}]%
        {henrich2010markets}
\bibfield{author}{\bibinfo{person}{Joseph Henrich}, \bibinfo{person}{Jean
  Ensminger}, \bibinfo{person}{Richard McElreath}, \bibinfo{person}{Abigail
  Barr}, \bibinfo{person}{Clark Barrett}, \bibinfo{person}{Alexander
  Bolyanatz}, \bibinfo{person}{Juan~Camilo Cardenas}, \bibinfo{person}{Michael
  Gurven}, \bibinfo{person}{Edwins Gwako}, \bibinfo{person}{Natalie Henrich},
  {et~al\mbox{.}}} \bibinfo{year}{2010}\natexlab{}.
\newblock \showarticletitle{Markets, religion, community size, and the
  evolution of fairness and punishment}.
\newblock \bibinfo{journal}{{\em science\/}} \bibinfo{volume}{327},
  \bibinfo{number}{5972} (\bibinfo{year}{2010}), \bibinfo{pages}{1480--1484}.
\newblock


\bibitem[\protect\citeauthoryear{Hochreiter and Schmidhuber}{Hochreiter and
  Schmidhuber}{1997}]%
        {Hochreiter:1997:LSM:1246443.1246450}
\bibfield{author}{\bibinfo{person}{Sepp Hochreiter} {and}
  \bibinfo{person}{J\"{u}rgen Schmidhuber}.} \bibinfo{year}{1997}\natexlab{}.
\newblock \showarticletitle{Long Short-Term Memory}.
\newblock \bibinfo{journal}{{\em Neural Comput.\/}} \bibinfo{volume}{9},
  \bibinfo{number}{8} (\bibinfo{date}{Nov.} \bibinfo{year}{1997}),
  \bibinfo{pages}{1735--1780}.
\newblock
\showISSN{0899-7667}
\showDOI{%
\url{https://doi.org/10.1162/neco.1997.9.8.1735}}


\bibitem[\protect\citeauthoryear{Jaderberg, Czarnecki, Dunning, Marris, Lever,
  Castaneda, Beattie, Rabinowitz, Morcos, Ruderman, et~al\mbox{.}}{Jaderberg
  et~al\mbox{.}}{2018}]%
        {jaderberg2018human}
\bibfield{author}{\bibinfo{person}{Max Jaderberg}, \bibinfo{person}{Wojciech~M
  Czarnecki}, \bibinfo{person}{Iain Dunning}, \bibinfo{person}{Luke Marris},
  \bibinfo{person}{Guy Lever}, \bibinfo{person}{Antonio~Garcia Castaneda},
  \bibinfo{person}{Charles Beattie}, \bibinfo{person}{Neil~C Rabinowitz},
  \bibinfo{person}{Ari~S Morcos}, \bibinfo{person}{Avraham Ruderman},
  {et~al\mbox{.}}} \bibinfo{year}{2018}\natexlab{}.
\newblock \showarticletitle{Human-level performance in first-person multiplayer
  games with population-based deep reinforcement learning}.
\newblock \bibinfo{journal}{{\em arXiv preprint arXiv:1807.01281\/}}
  (\bibinfo{year}{2018}).
\newblock


\bibitem[\protect\citeauthoryear{Jaques, Lazaridou, Hughes, Gulcehre, Ortega,
  Strouse, Leibo, and de~Freitas}{Jaques et~al\mbox{.}}{2018}]%
        {jaques2018intrinsic}
\bibfield{author}{\bibinfo{person}{Natasha Jaques}, \bibinfo{person}{Angeliki
  Lazaridou}, \bibinfo{person}{Edward Hughes}, \bibinfo{person}{Caglar
  Gulcehre}, \bibinfo{person}{Pedro~A Ortega}, \bibinfo{person}{DJ Strouse},
  \bibinfo{person}{Joel~Z Leibo}, {and} \bibinfo{person}{Nando de Freitas}.}
  \bibinfo{year}{2018}\natexlab{}.
\newblock \showarticletitle{Intrinsic Social Motivation via Causal Influence in
  Multi-Agent RL}.
\newblock \bibinfo{journal}{{\em arXiv preprint arXiv:1810.08647\/}}
  (\bibinfo{year}{2018}).
\newblock


\bibitem[\protect\citeauthoryear{Johnson and Gaines}{Johnson and
  Gaines}{1990}]%
        {doi:10.1146/annurev.es.21.110190.002313}
\bibfield{author}{\bibinfo{person}{M~L Johnson} {and} \bibinfo{person}{M~S
  Gaines}.} \bibinfo{year}{1990}\natexlab{}.
\newblock \showarticletitle{Evolution of Dispersal: Theoretical Models and
  Empirical Tests Using Birds and Mammals}.
\newblock \bibinfo{journal}{{\em Annual Review of Ecology and Systematics\/}}
  \bibinfo{volume}{21}, \bibinfo{number}{1} (\bibinfo{year}{1990}),
  \bibinfo{pages}{449--480}.
\newblock
\showDOI{%
\url{https://doi.org/10.1146/annurev.es.21.110190.002313}}
\showeprint{https://doi.org/10.1146/annurev.es.21.110190.002313}


\bibitem[\protect\citeauthoryear{Kempe, Lycett, and Mesoudi}{Kempe
  et~al\mbox{.}}{2014}]%
        {kempe2014cultural}
\bibfield{author}{\bibinfo{person}{Marius Kempe}, \bibinfo{person}{Stephen~J
  Lycett}, {and} \bibinfo{person}{Alex Mesoudi}.}
  \bibinfo{year}{2014}\natexlab{}.
\newblock \showarticletitle{From cultural traditions to cumulative culture:
  parameterizing the differences between human and nonhuman culture}.
\newblock \bibinfo{journal}{{\em Journal of theoretical biology\/}}
  \bibinfo{volume}{359} (\bibinfo{year}{2014}), \bibinfo{pages}{29--36}.
\newblock


\bibitem[\protect\citeauthoryear{Klyubin, Polani, and Nehaniv}{Klyubin
  et~al\mbox{.}}{2005}]%
        {klyubin2005empowerment}
\bibfield{author}{\bibinfo{person}{Alexander~S Klyubin},
  \bibinfo{person}{Daniel Polani}, {and} \bibinfo{person}{Chrystopher~L
  Nehaniv}.} \bibinfo{year}{2005}\natexlab{}.
\newblock \showarticletitle{Empowerment: A universal agent-centric measure of
  control}. In \bibinfo{booktitle}{{\em Evolutionary Computation, 2005. The
  2005 IEEE Congress on}}, Vol.~\bibinfo{volume}{1}. IEEE,
  \bibinfo{pages}{128--135}.
\newblock


\bibitem[\protect\citeauthoryear{Legg and Hutter}{Legg and Hutter}{2007}]%
        {legg2007universal}
\bibfield{author}{\bibinfo{person}{Shane Legg} {and} \bibinfo{person}{Marcus
  Hutter}.} \bibinfo{year}{2007}\natexlab{}.
\newblock \showarticletitle{Universal intelligence: A definition of machine
  intelligence}.
\newblock \bibinfo{journal}{{\em Minds and Machines\/}} \bibinfo{volume}{17},
  \bibinfo{number}{4} (\bibinfo{year}{2007}), \bibinfo{pages}{391--444}.
\newblock


\bibitem[\protect\citeauthoryear{Malthus}{Malthus}{1798}]%
        {malthus1798essay}
\bibfield{author}{\bibinfo{person}{Thomas~Robert Malthus}.}
  \bibinfo{year}{1798}\natexlab{}.
\newblock \bibinfo{booktitle}{{\em An essay on the principle of population: or,
  A view of its past and present effects on human happiness}}.
\newblock \bibinfo{publisher}{Reeves \& Turner}.
\newblock


\bibitem[\protect\citeauthoryear{Martin, Sasikumar, Everitt, and Hutter}{Martin
  et~al\mbox{.}}{2017}]%
        {martin2017count}
\bibfield{author}{\bibinfo{person}{Jarryd Martin},
  \bibinfo{person}{Suraj~Narayanan Sasikumar}, \bibinfo{person}{Tom Everitt},
  {and} \bibinfo{person}{Marcus Hutter}.} \bibinfo{year}{2017}\natexlab{}.
\newblock \showarticletitle{Count-based exploration in feature space for
  reinforcement learning}.
\newblock \bibinfo{journal}{{\em arXiv preprint arXiv:1706.08090\/}}
  (\bibinfo{year}{2017}).
\newblock


\bibitem[\protect\citeauthoryear{Mellars}{Mellars}{2006}]%
        {mellars2006did}
\bibfield{author}{\bibinfo{person}{Paul Mellars}.}
  \bibinfo{year}{2006}\natexlab{}.
\newblock \showarticletitle{Why did modern human populations disperse from
  Africa ca. 60,000 years ago? A new model}.
\newblock \bibinfo{journal}{{\em Proceedings of the National Academy of
  Sciences\/}} \bibinfo{volume}{103}, \bibinfo{number}{25}
  (\bibinfo{year}{2006}), \bibinfo{pages}{9381--9386}.
\newblock


\bibitem[\protect\citeauthoryear{Mnih, Badia, Mirza, Graves, Lillicrap, Harley,
  Silver, and Kavukcuoglu}{Mnih et~al\mbox{.}}{2016}]%
        {mnih2016asynchronous}
\bibfield{author}{\bibinfo{person}{Volodymyr Mnih},
  \bibinfo{person}{Adria~Puigdomenech Badia}, \bibinfo{person}{Mehdi Mirza},
  \bibinfo{person}{Alex Graves}, \bibinfo{person}{Timothy Lillicrap},
  \bibinfo{person}{Tim Harley}, \bibinfo{person}{David Silver}, {and}
  \bibinfo{person}{Koray Kavukcuoglu}.} \bibinfo{year}{2016}\natexlab{}.
\newblock \showarticletitle{Asynchronous methods for deep reinforcement
  learning}. In \bibinfo{booktitle}{{\em International conference on machine
  learning}}. \bibinfo{pages}{1928--1937}.
\newblock


\bibitem[\protect\citeauthoryear{Muthukrishna and Henrich}{Muthukrishna and
  Henrich}{2016}]%
        {muthukrishna2016innovation}
\bibfield{author}{\bibinfo{person}{Michael Muthukrishna} {and}
  \bibinfo{person}{Joseph Henrich}.} \bibinfo{year}{2016}\natexlab{}.
\newblock \showarticletitle{Innovation in the collective brain}.
\newblock \bibinfo{journal}{{\em Phil. Trans. R. Soc. B\/}}
  \bibinfo{volume}{371}, \bibinfo{number}{1690} (\bibinfo{year}{2016}),
  \bibinfo{pages}{20150192}.
\newblock


\bibitem[\protect\citeauthoryear{Ostrovski, Bellemare, Oord, and
  Munos}{Ostrovski et~al\mbox{.}}{2017}]%
        {ostrovski2017count}
\bibfield{author}{\bibinfo{person}{Georg Ostrovski}, \bibinfo{person}{Marc~G
  Bellemare}, \bibinfo{person}{Aaron van~den Oord}, {and}
  \bibinfo{person}{R{\'e}mi Munos}.} \bibinfo{year}{2017}\natexlab{}.
\newblock \showarticletitle{Count-based exploration with neural density
  models}.
\newblock \bibinfo{journal}{{\em arXiv preprint arXiv:1703.01310\/}}
  (\bibinfo{year}{2017}).
\newblock


\bibitem[\protect\citeauthoryear{Pathak, Agrawal, Efros, and Darrell}{Pathak
  et~al\mbox{.}}{2017}]%
        {pathak2017curiosity}
\bibfield{author}{\bibinfo{person}{Deepak Pathak}, \bibinfo{person}{Pulkit
  Agrawal}, \bibinfo{person}{Alexei~A Efros}, {and} \bibinfo{person}{Trevor
  Darrell}.} \bibinfo{year}{2017}\natexlab{}.
\newblock \showarticletitle{Curiosity-driven exploration by self-supervised
  prediction}. In \bibinfo{booktitle}{{\em International Conference on Machine
  Learning (ICML)}}, Vol.~\bibinfo{volume}{2017}.
\newblock


\bibitem[\protect\citeauthoryear{Powell, Shennan, and Thomas}{Powell
  et~al\mbox{.}}{2009}]%
        {powell2009late}
\bibfield{author}{\bibinfo{person}{Adam Powell}, \bibinfo{person}{Stephen
  Shennan}, {and} \bibinfo{person}{Mark~G Thomas}.}
  \bibinfo{year}{2009}\natexlab{}.
\newblock \showarticletitle{Late Pleistocene demography and the appearance of
  modern human behavior}.
\newblock \bibinfo{journal}{{\em Science\/}} \bibinfo{volume}{324},
  \bibinfo{number}{5932} (\bibinfo{year}{2009}), \bibinfo{pages}{1298--1301}.
\newblock


\bibitem[\protect\citeauthoryear{Richerson, Boyd, and Bettinger}{Richerson
  et~al\mbox{.}}{2009}]%
        {richerson2009cultural}
\bibfield{author}{\bibinfo{person}{Peter~J Richerson}, \bibinfo{person}{Robert
  Boyd}, {and} \bibinfo{person}{Robert~L Bettinger}.}
  \bibinfo{year}{2009}\natexlab{}.
\newblock \showarticletitle{Cultural innovations and demographic change}.
\newblock \bibinfo{journal}{{\em Human biology\/}} \bibinfo{volume}{81},
  \bibinfo{number}{3} (\bibinfo{year}{2009}), \bibinfo{pages}{211--235}.
\newblock


\bibitem[\protect\citeauthoryear{Schmidhuber}{Schmidhuber}{2010}]%
        {schmidhuber2010formal}
\bibfield{author}{\bibinfo{person}{J{\"u}rgen Schmidhuber}.}
  \bibinfo{year}{2010}\natexlab{}.
\newblock \showarticletitle{Formal theory of creativity, fun, and intrinsic
  motivation (1990--2010)}.
\newblock \bibinfo{journal}{{\em IEEE Transactions on Autonomous Mental
  Development\/}} \bibinfo{volume}{2}, \bibinfo{number}{3}
  (\bibinfo{year}{2010}), \bibinfo{pages}{230--247}.
\newblock


\bibitem[\protect\citeauthoryear{Silver, Huang, Maddison, Guez, Sifre, van~den
  Driessche, Schrittwieser, Antonoglou, Panneershelvam, Lanctot, Dieleman,
  Grewe, Nham, Kalchbrenner, Sutskever, Lillicrap, Leach, Kavukcuoglu, Graepel,
  and Hassabis}{Silver et~al\mbox{.}}{2016}]%
        {Silver16Go}
\bibfield{author}{\bibinfo{person}{David Silver}, \bibinfo{person}{Aja Huang},
  \bibinfo{person}{Chris~J. Maddison}, \bibinfo{person}{Arthur Guez},
  \bibinfo{person}{Laurent Sifre}, \bibinfo{person}{George van~den Driessche},
  \bibinfo{person}{Julian Schrittwieser}, \bibinfo{person}{Ioannis Antonoglou},
  \bibinfo{person}{Veda Panneershelvam}, \bibinfo{person}{Marc Lanctot},
  \bibinfo{person}{Sander Dieleman}, \bibinfo{person}{Dominik Grewe},
  \bibinfo{person}{John Nham}, \bibinfo{person}{Nal Kalchbrenner},
  \bibinfo{person}{Ilya Sutskever}, \bibinfo{person}{Timothy Lillicrap},
  \bibinfo{person}{Madeleine Leach}, \bibinfo{person}{Koray Kavukcuoglu},
  \bibinfo{person}{Thore Graepel}, {and} \bibinfo{person}{Demis Hassabis}.}
  \bibinfo{year}{2016}\natexlab{}.
\newblock \showarticletitle{Mastering the Game of {G}o with Deep Neural
  Networks and Tree Search.}
\newblock \bibinfo{journal}{{\em Nature\/}}  \bibinfo{volume}{529}
  (\bibinfo{year}{2016}), \bibinfo{pages}{484–--489}.
\newblock


\bibitem[\protect\citeauthoryear{Silver, Hubert, Schrittwieser, Antonoglou,
  Lai, Guez, Lanctot, Sifre, Kumaran, Graepel, Lillicrap, Simonyan, and
  Hassabis}{Silver et~al\mbox{.}}{2017}]%
        {silver2017chess}
\bibfield{author}{\bibinfo{person}{David Silver}, \bibinfo{person}{Thomas
  Hubert}, \bibinfo{person}{Julian Schrittwieser}, \bibinfo{person}{Ioannis
  Antonoglou}, \bibinfo{person}{Matthew Lai}, \bibinfo{person}{Arthur Guez},
  \bibinfo{person}{Marc Lanctot}, \bibinfo{person}{Laurent Sifre},
  \bibinfo{person}{Dharshan Kumaran}, \bibinfo{person}{Thore Graepel},
  \bibinfo{person}{Timothy Lillicrap}, \bibinfo{person}{Karen Simonyan}, {and}
  \bibinfo{person}{Demis Hassabis}.} \bibinfo{year}{2017}\natexlab{}.
\newblock \showarticletitle{Mastering chess and shogi by self-play with a
  general reinforcement learning algorithm}.
\newblock \bibinfo{journal}{{\em arXiv preprint arXiv:1712.01815\/}}
  (\bibinfo{year}{2017}).
\newblock


\bibitem[\protect\citeauthoryear{Smith}{Smith}{1776}]%
        {smith1776inquiry}
\bibfield{author}{\bibinfo{person}{Adam Smith}.}
  \bibinfo{year}{1776}\natexlab{}.
\newblock \bibinfo{booktitle}{{\em An inquiry into the nature and causes of the
  wealth of nations: Volume One}}.
\newblock \bibinfo{publisher}{London: printed for W. Strahan; and T. Cadell,
  1776.}
\newblock


\bibitem[\protect\citeauthoryear{Stewart and Stringer}{Stewart and
  Stringer}{2012}]%
        {stewart2012human}
\bibfield{author}{\bibinfo{person}{John~R Stewart} {and}
  \bibinfo{person}{Chris~B Stringer}.} \bibinfo{year}{2012}\natexlab{}.
\newblock \showarticletitle{Human evolution out of Africa: the role of refugia
  and climate change}.
\newblock \bibinfo{journal}{{\em Science\/}} \bibinfo{volume}{335},
  \bibinfo{number}{6074} (\bibinfo{year}{2012}), \bibinfo{pages}{1317--1321}.
\newblock


\bibitem[\protect\citeauthoryear{Tesauro}{Tesauro}{1995}]%
        {tesauro1995td}
\bibfield{author}{\bibinfo{person}{Gerald Tesauro}.}
  \bibinfo{year}{1995}\natexlab{}.
\newblock \showarticletitle{TD-Gammon, A Self-Teaching Backgammon Program,
  Achieves Master-Level Play}.
\newblock In \bibinfo{booktitle}{{\em Applications of Neural Networks}}.
  \bibinfo{publisher}{Springer}, \bibinfo{pages}{267--285}.
\newblock


\bibitem[\protect\citeauthoryear{Wade}{Wade}{1985}]%
        {doi:10.1086/284328}
\bibfield{author}{\bibinfo{person}{Michael~J. Wade}.}
  \bibinfo{year}{1985}\natexlab{}.
\newblock \showarticletitle{Soft Selection, Hard Selection, Kin Selection, and
  Group Selection}.
\newblock \bibinfo{journal}{{\em The American Naturalist\/}}
  \bibinfo{volume}{125}, \bibinfo{number}{1} (\bibinfo{year}{1985}),
  \bibinfo{pages}{61--73}.
\newblock
\showDOI{%
\url{https://doi.org/10.1086/284328}}
\showeprint{https://doi.org/10.1086/284328}


\bibitem[\protect\citeauthoryear{West, Pen, and Griffin}{West
  et~al\mbox{.}}{2002}]%
        {West72}
\bibfield{author}{\bibinfo{person}{Stuart~A. West}, \bibinfo{person}{Ido Pen},
  {and} \bibinfo{person}{Ashleigh~S. Griffin}.}
  \bibinfo{year}{2002}\natexlab{}.
\newblock \showarticletitle{Cooperation and Competition Between Relatives}.
\newblock \bibinfo{journal}{{\em Science\/}} \bibinfo{volume}{296},
  \bibinfo{number}{5565} (\bibinfo{year}{2002}), \bibinfo{pages}{72--75}.
\newblock
\showISSN{0036-8075}
\showDOI{%
\url{https://doi.org/10.1126/science.1065507}}
\showeprint{http://science.sciencemag.org/content/296/5565/72.full.pdf}


\bibitem[\protect\citeauthoryear{Yang, Yu, Bai, Wen, Zhang, and Wang}{Yang
  et~al\mbox{.}}{2018}]%
        {yang2018study}
\bibfield{author}{\bibinfo{person}{Yaodong Yang}, \bibinfo{person}{Lantao Yu},
  \bibinfo{person}{Yiwei Bai}, \bibinfo{person}{Ying Wen},
  \bibinfo{person}{Weinan Zhang}, {and} \bibinfo{person}{Jun Wang}.}
  \bibinfo{year}{2018}\natexlab{}.
\newblock \showarticletitle{A Study of AI Population Dynamics with
  Million-agent Reinforcement Learning}. In \bibinfo{booktitle}{{\em
  Proceedings of the 17th International Conference on Autonomous Agents and
  MultiAgent Systems}}. International Foundation for Autonomous Agents and
  Multiagent Systems, \bibinfo{pages}{2133--2135}.
\newblock


\end{thebibliography}

\end{document}